\documentclass{article}

\PassOptionsToPackage{numbers,sort&compress}{natbib}

\usepackage[preprint]{neurips_2026}

\usepackage[utf8]{inputenc}
\usepackage[T1]{fontenc}
\usepackage{url}
\usepackage{booktabs}
\usepackage{amsfonts}
\usepackage{amsmath}
\usepackage{amssymb}
\usepackage{nicefrac}
\usepackage{microtype}
\usepackage{xcolor}
\usepackage{graphicx}
\usepackage{multirow}
\usepackage{makecell}
\usepackage{tabularx}
\usepackage{enumitem}
\usepackage{array}
\usepackage{pifont}
\usepackage{placeins}
\usepackage[pagebackref,breaklinks,colorlinks]{hyperref}

\graphicspath{{figs_png/}}

\setlist[itemize]{topsep=2pt,itemsep=2pt,parsep=0pt,leftmargin=1.4em}

\newcommand{\degree}{\ensuremath{^\circ}}

\title{CD-RMOT-Bench: Benchmarking the Cross-Domain Referring Multi-Object Tracking}

\author{
	\begin{tabular}{c}
		Xiangqun Zhang$^{1}$
		\quad
		Likai Wang$^{1}$
		\quad
		Zekun Qian$^{1}$
		\\[-1pt]
		Ruize Han$^{2,*}$
		\quad
		Wei Feng$^{1}$
		\\[5pt]
		\textnormal{\small
			$^{1}$School of Computer Science and Technology,
			Tianjin University, Tianjin, China
		}
		\\
		\textnormal{\small
			$^{2}$Faculty of Computer Science and Artificial Intelligence,
		}
		\\[-1pt]
		\textnormal{\small
			Shenzhen University of Advanced Technology, Shenzhen, China
		}
		\\[4pt]
		\textnormal{\footnotesize
			\texttt{\{clzxq,kkww,clarkqian,wfeng\}@tju.edu.cn}
		}
		\\
		\textnormal{\footnotesize
			\texttt{hanruize@suat-sz.edu.cn}
		}
	\end{tabular}
}

\begin{document}
	
	\maketitle
	
	\begingroup
	\renewcommand{\thefootnote}{*}
	\footnotetext{Corresponding author.}
	\endgroup
	
	%	\vspace{-15pt}
	
	\begin{abstract}
		Referring multi-object tracking (RMOT) extends tracking from category-driven perception to language-guided understanding by grounding object trajectories in natural-language expressions.
		Despite recent progress, existing RMOT studies are largely conducted under in-domain settings, leaving the robustness of language-conditioned tracking under inevitable visual domain shifts unexplored.
		In this paper, we study \textbf{Cross-Domain Referring Multi-Object Tracking (CD-RMOT)}, a new and challenging problem that evaluates whether an RMOT model trained on a labeled source domain can reliably follow natural-language expressions in an unlabeled target domain with different visual conditions.
		To support systematic study, we construct \textbf{CD-RMOT-Bench}, a unified benchmark that combines real clear-domain referring tracking data, aligned digital-twin variants, and real adverse-domain videos.
		CD-RMOT-Bench enables both controlled weather/viewpoint shift analysis and realistic synthetic--real transfer evaluation under a shared RMOT protocol.
		Further, we provide a \textbf{Query-Centric Adaptation (QCA)} framework, designed to stabilize the query space that bridges visual trajectories and referring expressions.
		Extensive experiments reveal that domain shifts severely degrade RMOT performance, where the failure is not merely caused by object detection errors but more critically by unstable expression-conditioned temporal association and target selection.
		QCA establishes a strong baseline, while CD-RMOT-Bench opens a new direction for robust language-guided tracking across visual domains.
	\end{abstract}
	
	\section{Introduction}
	\label{sec:intro}
	
	\begin{figure}[t]
		\centering
		\includegraphics[width=0.9\linewidth]{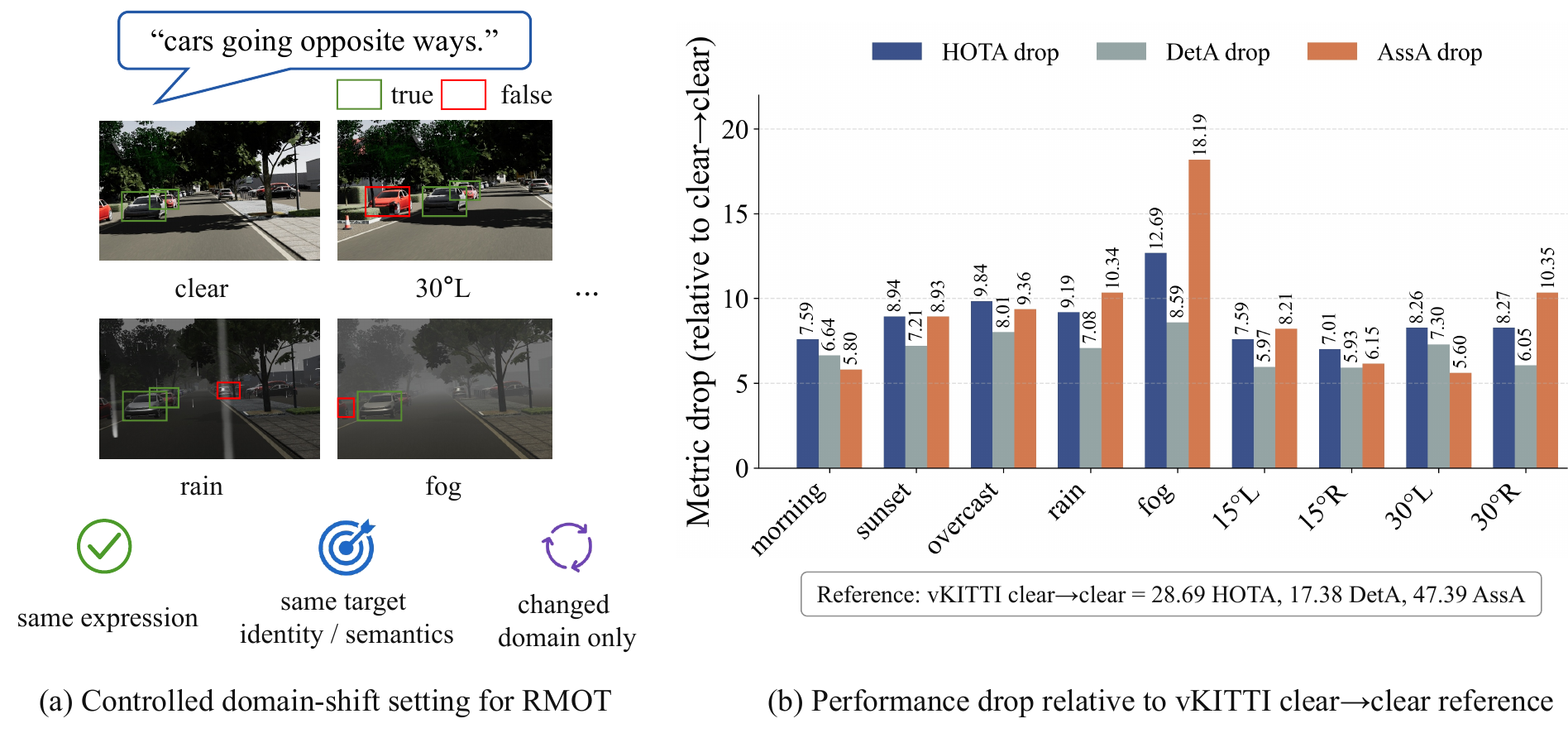}
		\vspace{-5pt}
		\caption{Controlled domain shift in RMOT. 
			(a) Under the same referring expression and target semantics, changing only the visual domain can cause the tracker to select expression-inconsistent trajectories. 
			(b) Existing RMOT model~\cite{zhang2024bootstrapping} shows clear HOTA, DetA, and AssA drops across weather and viewpoint shifts relative to the in-domain \textit{vKITTI clear}$\rightarrow$\textit{clear} reference.}
		\label{fig1}
		\vspace{-15pt}
	\end{figure}
	
	Multi-Object Tracking (MOT) is undergoing a paradigm shift from passive, category-based perception towards active, language-guided understanding, formalized as Referring Multi-Object Tracking. 
	Unlike traditional MOT, RMOT aims to simultaneously locate and track specific targets in video sequences based on natural language expressions, and has shown immense potential in human-computer interaction, Embodied AI, autonomous driving, \textit{etc.}, where intelligent agents must comprehend and seamlessly track specific targets driven by complex human intentions in highly dynamic environments.
	
	Recent RMOT benchmarks and methods have made encouraging progress under the standard in-domain setting, where training and testing videos are usually drawn from the same visual distribution.
	However, when deployed in real-world scenarios, perception systems inevitably encounter severe domain shifts, including extreme weather variations, unpredictable sensor viewpoint changes, and inherent synthetic-to-real realism gaps. 
	As shown in Fig.~\ref{fig1}, such shifts pose a critical challenge to existing RMOT models.
	Fig.~\ref{fig1}(a) presents a controlled case where the referring expression and target semantics remain unchanged, while only the visual domain varies across weather and viewpoint conditions. Although the model can still detect visually plausible objects, its selected trajectories may deviate from the language-specified targets.
	Fig.~\ref{fig1}(b) further quantifies this effect, showing consistent performance drops relative to the in-domain reference.
	These observations indicate that current RMOT evaluation largely overlooks a fundamental question: \textit{can a language-conditioned tracker remain reliable when the visual domain changes}?

	To address this gap, we introduce a novel and challenging problem, Cross-Domain Referring Multi-Object Tracking, which evaluates whether an RMOT model trained in a labeled source domain can generalize to an unlabeled target domain with different visual conditions.
	The core challenge of CD-RMOT lies in maintaining the semantic fidelity of language-guided tracking trajectories when the underlying visual distribution shifts dramatically, without access to target-domain tracking annotations.

	A principled benchmark for this new problem is difficult to build.
	Real-world videos naturally contain domain variations, but they also change scene layout, object population, camera motion, and target interactions, making it hard to determine whether failure comes from domain shift or from different semantic content.
	Conversely, purely synthetic benchmarks can provide controllable variations, but they may not reflect the realism gap faced by practical tracking systems.
	What is needed is a benchmark that \textit{combines controlled domain shift with realistic cross-domain transfer}, so that models can be evaluated along both axes.
	To this end, we introduce CD-RMOT-Bench, a unified benchmark for CD-RMOT, which integrates real-world referring annotations, aligned digital-twin variants, and adverse real-domain videos into a single evaluation suite.
	Its controlled part enables weather and viewpoint changes to be studied while preserving the underlying scene structure and referring semantics, while its real-domain part evaluates generalization across synthetic--real domains.

	Further, we propose a Query-Centric Adaptation framework for CD-RMOT.
	The method is motivated by the observation that domain shift can destabilize language-conditioned temporal queries, causing representation mismatch and semantic drift during target selection.
	To address this issue, we introduce temporal teacher-student consistency to stabilize target-domain queries, adaptive domain adversarial learning to align transferable query representations, and language-guided adaptation to preserve the semantic correspondence between queries and referring expressions.
	Extensive experiments on CD-RMOT-Bench reveal critical insights: visual domain shifts severely degrade expression-conditioned tracking, and this degradation is driven primarily by temporal association and selection failures rather than mere detection errors. The proposed QCA significantly recovers these performance drops across weather, viewpoint, and synthetic--real transfer settings, establishing a strong and reliable baseline.
	
	The main contributions are summarized as follows:
	\begin{itemize}
		\item We pioneer the problem of \textbf{Cross-Domain Referring Multi-Object Tracking}, highlighting the critical need for robust visual-language tracking under domain shifts.
		\item We construct \textbf{CD-RMOT-Bench}, a unified CD-RMOT benchmark covering controlled weather and viewpoint variations and realistic synthetic--real domain transfer under a shared evaluation protocol. Its controlled core, Refer-vKITTI, is built through per-scene identity mapping, domain-specific expression-track transfer, and residual quality control.
		\item We provide a query-centric adaptation method that improves CD-RMOT through reliable query stabilization, task-relevant query alignment, and language-anchored query calibration.
		Extensive experiments demonstrate the severe limitations of existing methods under domain shifts and validate the effectiveness of the proposed framework as a strong baseline for future research.
	\end{itemize}
	
	\section{Related Work}
	\label{sec:rela}
	
	\paragraph{Referring Multi-Object Tracking.}
	
	Language-guided tracking extends object tracking by selecting targets through natural-language expressions. Modern MOT has developed strong detection, association, and temporal reasoning mechanisms, from online tracking-by-detection methods~\cite{bewley2016sort,wojke2017simple,bergmann2019tracking,zhou2020tracking,zhang2021fairmot,zhang2022bytetrack,cao2023ocsort} to transformer/query-based trackers~\cite{sun2020transtrack,meinhardt2022trackformer,zeng2022motr,zhang2023motrv2,yan2022unicorn}. These methods provide useful tracking foundations, but they do not determine trajectories from open-ended referring expressions.
	
	Referring multi-object tracking requires the model to track only the objects specified by a natural-language query. TransRMOT~\cite{wu2023rmot} establishes the RMOT task and benchmark, and TempRMOT~\cite{zhang2024bootstrapping} expands linguistic difficulty through Refer-KITTI-V2. Recent RMOT studies further improve language fusion, plug-and-play referring control, multimodal cues, fine-grained semantic guidance, and reasoning-oriented target selection, including iKUN~\cite{du2024ikun}, EchoTrack~\cite{lin2024echotrack}, DeepRMOT~\cite{he2024deeprmot}, MGLT~\cite{chen2025multigranularity}, DKGTrack~\cite{li2025dkgtrack}, and recent reasoning or zero-shot variants~\cite{chen2025reamot,chamiti2025refergpt}. These works mainly evaluate closed or nearly in-domain behavior, leaving cross-domain robustness of expression-conditioned trajectories underexplored.
	
	A related line is referring video object segmentation (RVOS), where language guides temporally grounded mask prediction. Representative methods include MTTR~\cite{botach2022end}, ReferFormer~\cite{wu2022language}, OnlineRefer~\cite{wu2023onlinerefer}, MUTR~\cite{yan2024referred}, and ReferDINO~\cite{liang2025referdino}. RVOS provides useful cross-task references, and we adapt MUTR and ReferDINO as baselines. Yet RVOS outputs masks rather than identity-aware multi-object trajectories, and its standard protocol does not diagnose tracking association or referring consistency under domain shift.
	
	%	\vspace{-5pt}
	
	\paragraph{Cross-Domain Adaptation.}
	
	Cross-domain adaptation has been widely studied for recognition, detection, and structured prediction. General adaptation methods reduce distribution mismatch through discrepancy minimization or adversarial learning, including DAN~\cite{long2015learning}, DANN~\cite{ganin2016dann}, ADDA~\cite{tzeng2017adversarial}, CDAN~\cite{long2018conditional}, and MCD~\cite{saito2018maximum}. For object detection, DA-Faster R-CNN~\cite{chen2018dafaster}, SWDA~\cite{saito2019swda}, MTOR~\cite{cai2019exploring}, Unbiased Mean Teacher~\cite{deng2021unbiased}, Adaptive Teacher~\cite{li2022adaptive}, Probabilistic Teacher~\cite{chen2022probabilistic}, and recent prompt-based adaptation methods~\cite{zhang2025upre} show that instance-level alignment and teacher--student training can improve transfer under unlabeled target domains.
	
	Video and test-time adaptation further address temporal and non-stationary shifts. TA$^3$N~\cite{chen2019temporal} studies temporal attentive alignment for video domain adaptation, while Tent~\cite{wang2021tent}, CoTTA~\cite{wang2022cotta}, EATA~\cite{niu2022eata}, PETAL~\cite{brahma2023petal}, and related synthetic-to-real video studies~\cite{zhang2025synthetic,wang2025new} show that adaptation remains challenging beyond static images. These works motivate target-query consistency, query-level alignment, and language-guided correction in our reference method. However, they do not define a benchmark protocol for expression-conditioned multi-object trajectories, where detection, association, and language-conditioned target selection must be evaluated jointly.
	
	%	\vspace{-5pt}
	
	\paragraph{Benchmarks for Visual Domain Shift.}
	
	Driving and tracking benchmarks such as KITTI~\cite{geiger2012kitti}, MOTChallenge~\cite{milan2016mot16}, Cityscapes~\cite{cordts2016cityscapes}, BDD100K~\cite{yu2020bdd100k}, and nuScenes~\cite{caesar2020nuscenes} have shaped progress in perception, but they do not provide controlled variants where scene semantics and referring expressions remain fixed while the visual domain changes. Synthetic and digital-twin datasets, including Virtual KITTI~\cite{gaidon2016virtual}, Virtual KITTI 2~\cite{cabon2020vkitti2}, SYNTHIA~\cite{ros2016synthia}, GTA5~\cite{richter2016playing}, and SHIFT~\cite{sun2022shift}, support structured domain variation, but are not designed for referring-expression-conditioned tracking. General domain-shift benchmarks such as VisDA~\cite{peng2017visda}, DomainNet~\cite{peng2019moment}, and WILDS~\cite{koh2021wilds} similarly do not evaluate language-conditioned multi-object trajectories.
	
	Existing RMOT datasets provide language-conditioned trajectories, while controlled synthetic datasets provide domain variation. Recent extensions broaden RMOT-style evaluation toward reasoning-intensive or aerial scenarios~\cite{chen2025reamot,chen2026aerialmind}, but they do not provide controlled paired domain shifts for driving RMOT. CD-RMOT-Bench connects these two sides by transferring Refer-KITTI-V2 expressions to aligned Virtual KITTI 2 digital-twin scenes and combining the resulting controlled core with real-domain anchors.
	
	\section{CD-RMOT-Bench Overview}
	\label{sec:benchmark}

	\subsection{Motivation and Design}
	
	CD-RMOT-Bench is designed as a comprehensive testbed for language-conditioned UDA in RMOT. In this setting, target-domain videos and referring expressions are available as task inputs during adaptation, but target-domain boxes, track identities, masks, and expression-object associations are reserved for evaluation only. This distinction is important: target expressions specify what should be tracked, but they do not reveal the target-domain answer.
	
	Evaluating domain shift in dynamic video environments is difficult because visual changes are often entangled with changes in scene layout, object population, camera motion, target interaction, and expression semantics. CD-RMOT-Bench therefore evaluates CD-RMOT along two complementary axes. The first is \textbf{controlled domain shift}, where weather and viewpoint changes can be studied under shared scene structure and stable referring semantics. The second is \textbf{realistic cross-domain transfer}, where models are evaluated across synthetic--real and clear--adverse domain gaps that better reflect deployment conditions.
	
	CD-RMOT-Bench addresses these axes by organizing three complementary resources into a unified benchmark family. Refer-KITTI-V2~\cite{zhang2024bootstrapping} serves as the real clear-domain anchor and provides real-domain referring expressions and trajectories. Refer-vKITTI forms the controlled digital-twin core, where aligned Virtual KITTI 2~\cite{cabon2020vkitti2} variants isolate weather and viewpoint shifts while preserving scene structure as much as possible. Refer-BDD~\cite{lin2024echotrack} provides a real adverse-domain anchor that complements the controlled setting with realistic domain variation. Together, these resources enable controlled weather/viewpoint analysis and bidirectional synthetic--real evaluation under a shared RMOT protocol.
	
	\subsection{Benchmark Construction}
	
	%	\begin{figure}[t]
		\begin{figure}[!htbp]
			\centering
			\includegraphics[width=0.88\linewidth]{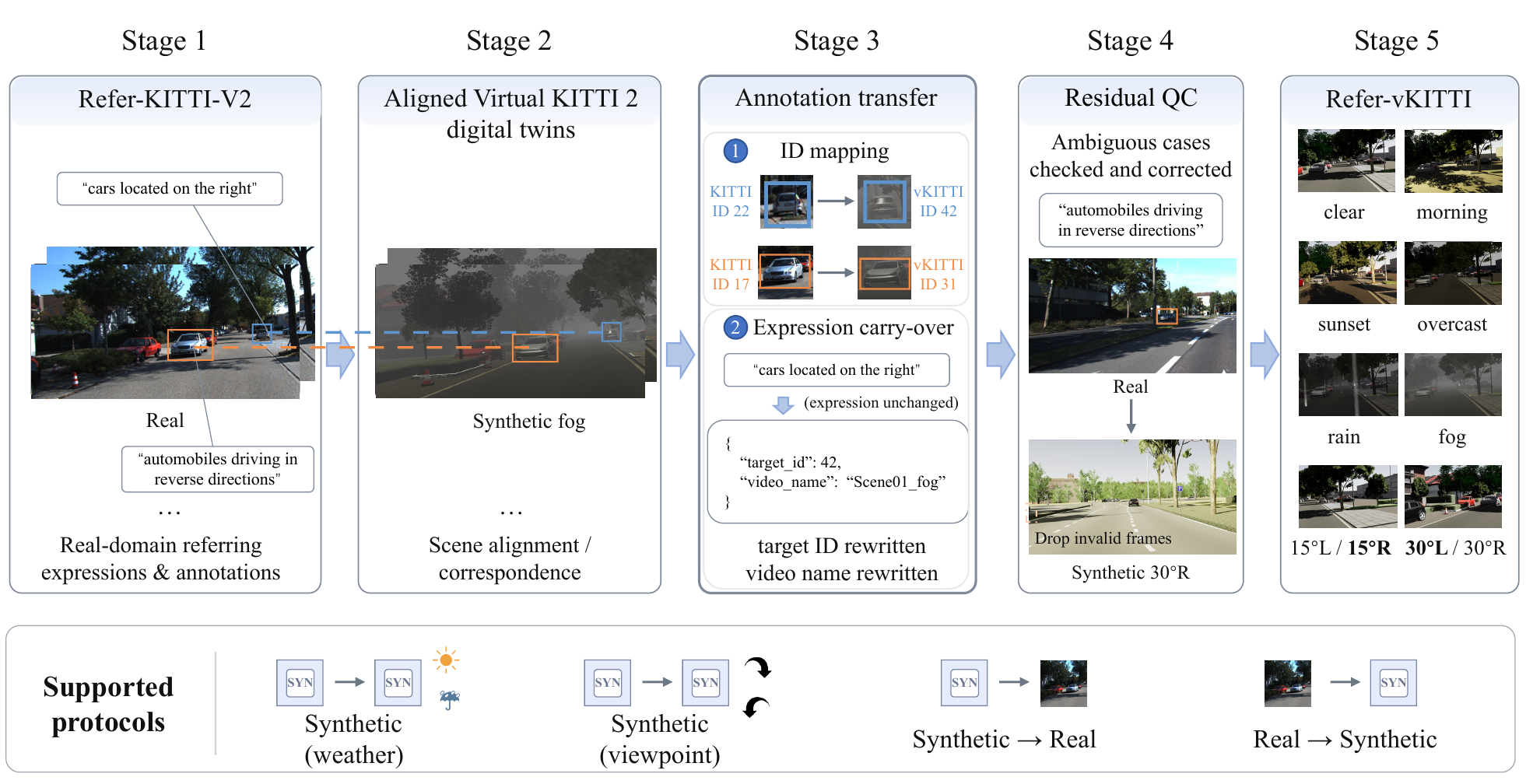}
			\caption{Refer-vKITTI construction and supported protocols. Refer-KITTI-V2 provides real-domain expressions and annotations, while aligned Virtual KITTI 2 scenes provide controlled digital-twin domains. We first establish per-scene identity mappings, then transfer expression-track annotations by preserving descriptions and rewriting target identities and video names. Residual quality control handles ambiguous correspondences and visibility cases. Refer-vKITTI supports weather/viewpoint transfer and, with Refer-KITTI-V2 and Refer-BDD, bidirectional synthetic--real evaluation.}
			\label{fig:benchmark_construction}
			\vspace{-10pt}
		\end{figure}
	
	Fig.~\ref{fig:benchmark_construction} shows how CD-RMOT-Bench realizes the two axes introduced above. Stages 1--5 construct Refer-vKITTI as the controlled digital-twin core for weather/viewpoint shift analysis, while the bottom protocol row connects it with Refer-KITTI-V2 and Refer-BDD for synthetic--real transfer evaluation.
	
	\textbf{Stage 1--2: source annotations and aligned digital twins.}
	We start from Refer-KITTI-V2~\cite{zhang2024bootstrapping}, which provides real driving scenes with referring expressions and expression-conditioned trajectories. We then use aligned Virtual KITTI 2 scenes~\cite{cabon2020vkitti2}, whose digital-twin structure provides controlled weather and viewpoint variants of KITTI-like scenes. These two stages implement the \emph{controlled domain shift} axis: real-domain expressions and trajectories are carried to aligned synthetic domains where weather or viewpoint can change while the scene structure remains comparable.
	
	\textbf{Stage 3: mapping-based annotation transfer.}
	The core construction follows a mapping-then-transfer procedure. For each aligned real/synthetic scene, we first establish per-scene KITTI-to-vKITTI identity correspondences. Candidate correspondences are supported by frame-level IoU evidence and aggregated into track-level scores that incorporate geometric consistency, area similarity, and temporal overlap. The final one-to-one correspondences provide an auditable identity mapping between real and synthetic scenes. Using this mapping, we transfer Refer-KITTI-V2 expression-track annotations to vKITTI domains: the natural-language expression is preserved, while the target identity and video name are rewritten according to the mapped vKITTI identity and target domain. Each mapped expression-track pair is instantiated across \textit{clear}, \textit{rain}, \textit{fog}, \textit{morning}, \textit{overcast}, \textit{sunset}, \textit{15\degree L}, \textit{15\degree R}, \textit{30\degree L}, and \textit{30\degree R}, where L/R denote left/right viewpoint shifts.
	
	\textbf{Stage 4--5: residual quality control and Refer-vKITTI.}
	After automatic transfer, residual quality control handles ambiguous cases, including weak correspondence evidence, identity ambiguity, expression--target inconsistency, localization anomalies, and visibility-induced invalidations. Frame-level annotations are retained only when the mapped target is present in the corresponding vKITTI labels. The resulting dataset, Refer-vKITTI, forms the controlled digital-twin core of CD-RMOT-Bench, supporting synthetic weather and viewpoint transfer. Together with Refer-KITTI-V2 and Refer-BDD, it further enables synthetic-to-real and real-to-synthetic evaluation, as summarized in the bottom protocol row of Fig.~\ref{fig:benchmark_construction}.
	
	\textbf{Real adverse-domain anchor.}
	To support the \emph{realistic cross-domain transfer} axis, Refer-BDD provides a BDD100K-based referring tracking set following recent AR-MOT/RMOT evaluation practice~\cite{lin2024echotrack,chen2025multigranularity}. Built from the BDD100K tracking source~\cite{yu2020bdd100k}, it is not part of the digital-twin transfer pipeline. Instead, it serves as a complementary real adverse-domain anchor for realism-shift evaluation. To enable direct comparison, Refer-KITTI-V2, Refer-vKITTI, and Refer-BDD are normalized into a shared evaluation structure, where each expression is represented by a sentence and a frame-indexed set of target track identities.
	
	\textbf{Construction quality.}
	The automatic pipeline yields 285{,}160 initially transferred expression--track pairs after expanding five aligned Refer-KITTI-V2 sequences to 10 vKITTI domains. After visibility filtering and format normalization, 272{,}043 pairs are retained, corresponding to a 95.40\% retention rate. At the identity-mapping level, 6{,}887 temporally overlapping source--target identity candidates are reduced to 374 IoU-supported candidates, and the final scene-level mapping contains 251 matched identity correspondences reused across the 10 vKITTI domains. Detailed scoring and construction statistics are provided in \textbf{Appendix~\ref{sec:appendix_construction}}.
	
	\subsection{Statistics and Evaluation Protocol}
	
	Table~\ref{tab:benchmark_overview} reports the scale, role, and domain coverage of each subset and the overall benchmark.
	
	\begin{table}[!htbp]
		\caption{Overview of CD-RMOT-Bench. Refer-KITTI-V2, Refer-vKITTI, and Refer-BDD respectively serve as the real clear-domain anchor, controlled digital-twin core, and real adverse-domain anchor for unified CD-RMOT evaluation.}
		\centering
		\scriptsize
		\renewcommand{\arraystretch}{1.12}
		\begin{tabular*}{0.9\linewidth}{@{\extracolsep{\fill}} l c r r r r r l @{}}
			\toprule
			Dataset & Type & Videos & Frames & Exprs. & Expr./Video & Words/Expr. & Coverage \\
			\midrule
			Refer-KITTI-V2~\cite{zhang2024bootstrapping} & Real & 21 & 8{,}008 & 9{,}778 & 465.62 & 6.64 & Real clear scenes \\
			Refer-vKITTI & Synthetic & 50 & 20{,}560 & 27{,}730 & 554.60 & 7.26 & 10 controlled domains \\
			Refer-BDD~\cite{lin2024echotrack} & Real & 50 & 10{,}005 & 4{,}610 & 92.20 & 6.30 & Real adverse scenes \\
			\midrule
			CD-RMOT-Bench & Mixed & 121 & 38{,}573 & 42{,}118 & 348.08 & 6.91 & \makecell[l]{Weather/viewpoint;\\synthetic--real transfer} \\
			\bottomrule
		\end{tabular*}
		\label{tab:benchmark_overview}
	\end{table}
	
	\begin{figure}[t]
		\centering
		\includegraphics[width=0.8\linewidth]{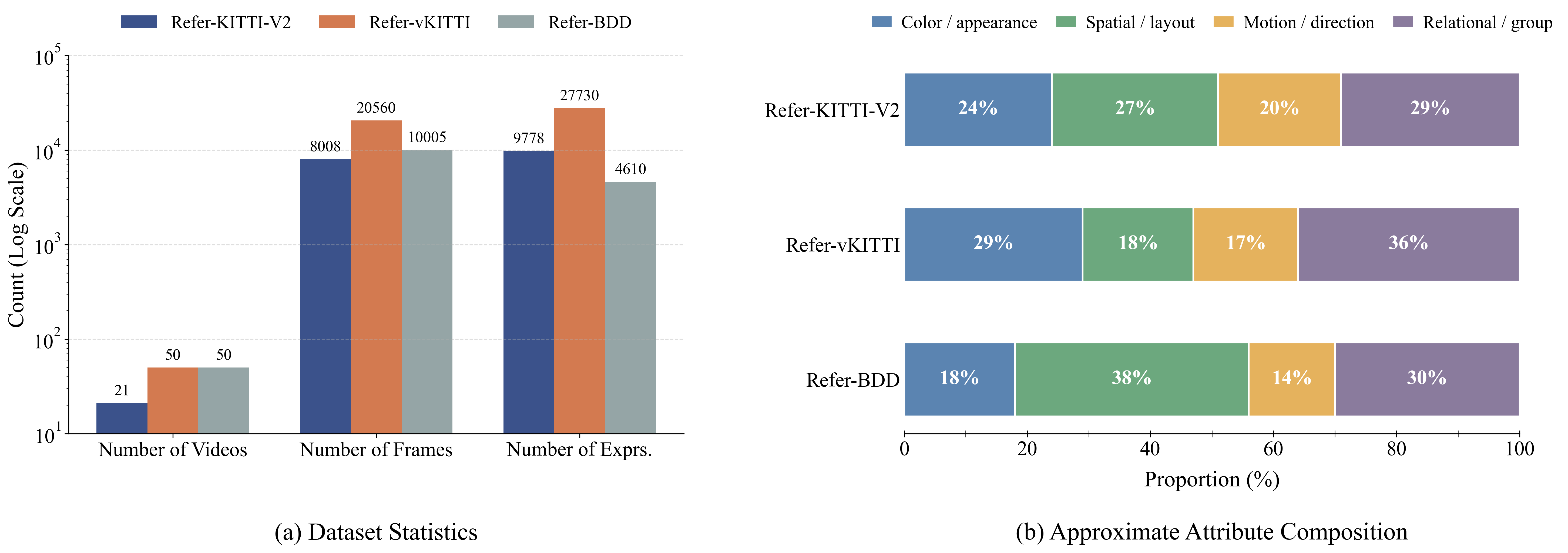}
		\caption{Benchmark characterization of CD-RMOT-Bench. Refer-vKITTI provides the controlled digital-twin core, while Refer-KITTI-V2 and Refer-BDD serve as real-domain anchors. Expression-cue proportions are normalized over coarse keyword counts for visualization; categories are non-exclusive and used only for characterization, not as additional manual labels.}
		\label{fig:benchmark_characterization}
		\vspace{-5pt}
	\end{figure}

	\textbf{Scale and expression cues.}
	As shown in Table~\ref{tab:benchmark_overview}, CD-RMOT-Bench contains 121 videos, 38{,}573 frames, and 42{,}118 expressions. The three subsets serve complementary roles: Refer-KITTI-V2 provides the real clear-domain anchor, Refer-vKITTI forms the controlled digital-twin core, and Refer-BDD contributes the real adverse-domain anchor. In particular, Refer-vKITTI contains 50 videos across 10 controlled domains, including \textit{clear}, \textit{fog}, \textit{rain}, \textit{morning}, \textit{overcast}, \textit{sunset}, \textit{15\degree L}, \textit{15\degree R}, \textit{30\degree L}, and \textit{30\degree R}, supporting controlled evaluation without conflating visual shift with unrelated scene changes.
	
	Fig.~\ref{fig:benchmark_characterization} characterizes the benchmark composition and expression cues. Fig.~\ref{fig:benchmark_characterization}(a) shows the composition of Refer-KITTI-V2, Refer-vKITTI, and Refer-BDD across real clear-domain, controlled digital-twin, and real adverse-domain settings. Fig.~\ref{fig:benchmark_characterization}(b) reports coarse expression-cue statistics over four non-exclusive categories: \emph{color/appearance}, \emph{spatial/layout}, \emph{motion/direction}, and \emph{relational/group}. Here, relational/group cues include object--object relations, scene-relative references, ordinal descriptions, and group-level target semantics. The distribution shows that CD-RMOT-Bench covers diverse referring evidence rather than only simple appearance-based descriptions.
	
	\textbf{Evaluation protocols.}
	CD-RMOT-Bench supports four evaluation modes under a shared RMOT protocol. \textbf{Weather transfer} evaluates changes from \textit{vKITTI clear} to weather variants such as \textit{fog} and \textit{rain}, isolating appearance and visibility changes under shared scene semantics. \textbf{Viewpoint transfer} evaluates geometric sensitivity from \textit{vKITTI clear} to shifted camera viewpoints such as \textit{30\degree L}. \textbf{Synthetic-to-real transfer} evaluates realism gaps from virtual scenes to Refer-KITTI-V2 or Refer-BDD. \textbf{Real-to-synthetic transfer} provides the reverse direction by using real clear-domain data as source and controlled synthetic domains as targets. Together, Refer-vKITTI supports controlled weather/viewpoint analysis, while Refer-KITTI-V2 and Refer-BDD complement it with cross-boundary synthetic--real and clear--adverse evaluation. Detailed protocol splits, frame counts, and expression counts are provided in \textbf{Appendix~\ref{sec:appendix_protocol}}.
	
	\section{QCA: A Query-Centric Adaptation Reference Method}
	\label{sec:method}
	
	\subsection{Overview}
	
	QCA is used as a query-centric reference method for probing the gap exposed by CD-RMOT-Bench. The benchmark diagnosis suggests that cross-domain degradation is not only caused by missing detections. Under domain shift, visually plausible candidates may remain available, while the language-conditioned temporal queries that determine the final trajectories become unstable, mismatched across domains, or semantically drifted from the referring expression. QCA therefore operates on temporal object queries produced by a shared RMOT backbone, while the overall setting remains language-conditioned UDA for RMOT.
	
	Let $\mathcal{D}_s=\{(V_s,E_s,Y_s)\}$ denote the labeled source-domain RMOT training set, where $V_s$ is a video clip, $E_s$ is the referring expression, and $Y_s$ contains expression-conditioned tracking annotations. Let $\mathcal{D}_t=\{(V_t,E_t)\}$ denote the target-domain adaptation set, where target videos and referring expressions are available as task inputs but target boxes, identities, masks, and expression-object associations are not used as supervised labels. Given a video clip and a referring expression, the shared RMOT backbone outputs a set of language-conditioned temporal queries
	\begin{equation}
		Q=\{q_1,\dots,q_N\}, \qquad q_i\in\mathbb{R}^d,
	\end{equation}
	where each query is associated with an objectness logit $o_i$, a box prediction $b_i$, and a referring logit $r_i$.
	
\begin{figure}[t]
		\centering
		\includegraphics[width=0.9\linewidth]{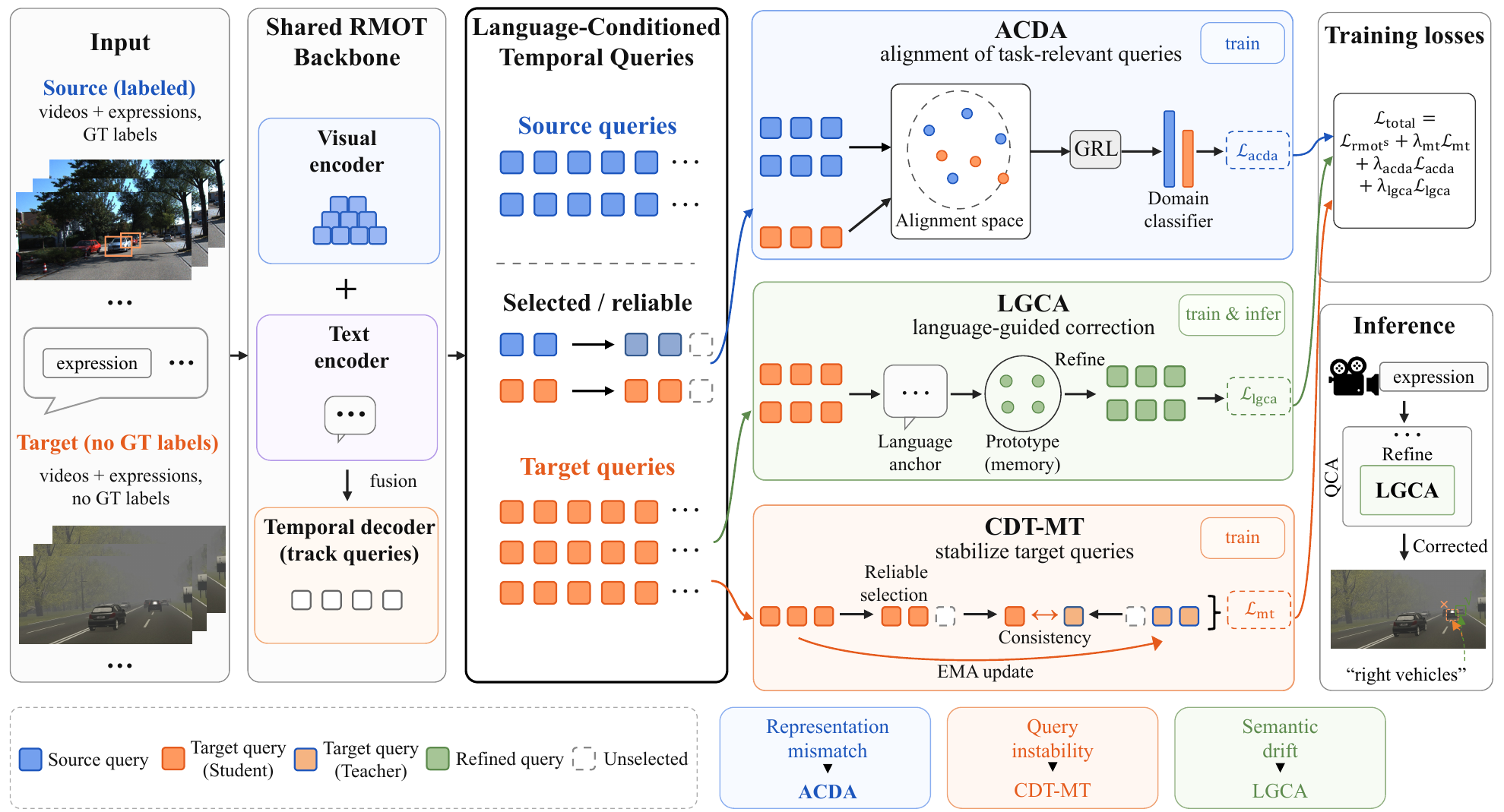}
		\caption{Overview of QCA as a query-centric reference method for language-conditioned UDA in RMOT. The shared RMOT backbone produces language-conditioned temporal queries, while CDT-MT, ACDA, and LGCA target three failure modes exposed by CD-RMOT-Bench: target-query instability, task-relevant source--target mismatch, and language-conditioned semantic drift.}
		\label{fig2}
	\end{figure}

	As illustrated in Fig.~\ref{fig2}, QCA adds three query-level adaptation mechanisms on top of the supervised source-domain RMOT objective. Cross-Domain Temporal Mean Teacher (CDT-MT) stabilizes reliable target-domain queries through teacher--student consistency. Adaptive Curricular Domain Adversarial (ACDA) aligns task-relevant source and target queries using naturally available domain labels from the source/target split. Language-Guided Continuous Adaptation (LGCA) uses the referring expression as a semantic anchor to reduce query drift and preserve expression-consistent selection. The overall objective is
	\begin{equation}
		\mathcal{L}_{\text{total}}
		=
		\mathcal{L}_{\text{rmot}}^{s}
		+
		\lambda_{\text{mt}}\mathcal{L}_{\text{mt}}
		+
		\lambda_{\text{acda}}\mathcal{L}_{\text{acda}}
		+
		\lambda_{\text{lgca}}\mathcal{L}_{\text{lgca}},
		\label{eq:overall_loss}
	\end{equation}
	where $\mathcal{L}_{\text{rmot}}^{s}$ is the supervised RMOT loss on the source domain, and $\lambda_{\text{mt}},\lambda_{\text{acda}},\lambda_{\text{lgca}}$ are trade-off weights.
	
	\vspace{-10pt}
	
	\subsection{Query-Centric Adaptation Learning}
	
	CDT-MT addresses target-query instability. Under domain shift, a query may remain associated with a plausible object but fluctuate in feature representation, box prediction, or referring confidence across training updates. We maintain a student model $f_\theta$ and an EMA teacher $f_{\bar{\theta}}$, updated as
	\begin{equation}
		\bar{\theta}^{(k+1)}
		=
		\alpha \bar{\theta}^{(k)}
		+
		(1-\alpha)\theta^{(k)},
		\label{eq:ema}
	\end{equation}
	and optimize selective consistency over reliable target queries:
	\begin{equation}
		\mathcal{L}_{\text{mt}}
		=
		\sum_{i\in\Omega_t} w_i \Bigl(
		\lambda_f \|q_i^{\text{stu}}-q_i^{\text{tea}}\|_2^2
		+
		\lambda_b \,\ell_{\text{smooth-}L_1}(b_i^{\text{stu}},b_i^{\text{tea}})
		+
		\lambda_r \|\sigma(r_i^{\text{stu}})-\sigma(r_i^{\text{tea}})\|_2^2
		\Bigr),
		\label{eq:mt_loss}
	\end{equation}
	where $\Omega_t$ denotes the retained reliable subset and $w_i$ its associated soft weights.
	
	ACDA addresses task-relevant source--target mismatch at the query level. Instead of aligning global scene features, it focuses on object queries that are likely to matter for RMOT. The source/target split naturally provides domain labels, so this adversarial signal does not require additional manual domain annotation. For query $i$, the task-relevance score is
	\begin{equation}
		c_i = \sigma(o_i)\cdot \sigma(r_i),
		\label{eq:acda_task_conf}
	\end{equation}
	and the resulting adversarial objective is
	\begin{equation}
		\mathcal{L}_{\text{acda}}
		=
		\frac{1}{M}
		\sum_{i=1}^{M}
		w_i^{\text{acda}}
		\,
		\ell_{\text{ce}}
		\bigl(
		g(\mathrm{GRL}(z_i)), d_i
		\bigr),
		\label{eq:acda_loss}
	\end{equation}
	where $\{z_i\}_{i=1}^{M}$ are the retained task-relevant query features, $g(\cdot)$ is the domain classifier, $d_i$ is the source/target domain label, and $w_i^{\text{acda}}$ is the curriculum weight.
	
	LGCA addresses language-conditioned semantic drift. Even when detections remain plausible, the selected query may move away from the referring expression over time. Given query embeddings $\{q_i\}_{i=1}^{N}$, tracking confidences $\{\nu_i\}_{i=1}^{N}$, and a pooled text anchor $\ell\in\mathbb{R}^{d}$, we define language-guided confidence as
	\begin{equation}
		\kappa_i = \frac{1+\cos(q_i,\ell)}{2}\cdot \nu_i.
		\label{eq:lgca_conf}
	\end{equation}
	Let $\mathcal{P}$ denote the selected confident subset. The current prototype estimate and its momentum update are
	\begin{equation}
		\hat{p}
		=
		\sum_{i\in\mathcal{P}}
		\bar{\kappa}_i q_i,
		\qquad
		\bar{\kappa}_i=\frac{\kappa_i}{\sum_{j\in\mathcal{P}}\kappa_j+\varepsilon},
		\label{eq:lgca_proto_est}
	\end{equation}
	\begin{equation}
		p^{(n)} = m\,p^{(n-1)} + (1-m)\hat{p},
		\label{eq:lgca_proto_update}
	\end{equation}
	and apply the residual refinement
	\begin{equation}
		q_i^{\text{out}}
		=
		(1-\alpha_i)q_i + \alpha_i q_i^{*},
		\label{eq:lgca_blend}
	\end{equation}
	where $q_i^{*}$ denotes the refined query feature and $\alpha_i$ is modulated by the current quality score. In this way, LGCA uses language not as an additional label source, but as a semantic anchor for maintaining expression-consistent temporal selection. Additional module details, schedules, and hyperparameters are provided in \textbf{Appendix~\ref{sec:appendix_method}}.
	
	\section{Experiments}
	\label{sec:exp}
	
	\subsection{Experimental Setup}
	
	We instantiate the protocols defined in Sec.~\ref{sec:benchmark} under a unified TrackEval-based RMOT evaluation setting. The main experiments cover weather transfer from \textit{vKITTI clear} to \textit{fog} and \textit{rain}, viewpoint transfer from \textit{vKITTI clear} to \textit{30\degree L}, synthetic-to-real transfer to Refer-KITTI-V2 and Refer-BDD, and the reverse real-to-synthetic direction from Refer-KITTI-V2 to \textit{vKITTI fog}. All settings use fixed protocol splits.
	
	All experiments follow language-conditioned UDA for RMOT. Source-domain videos, referring expressions, and tracking annotations are used for supervised RMOT training. Target-domain videos and referring expressions are used as task-conditioned inputs for adaptation, but target-domain bounding boxes, track identities, masks, and expression-object associations are not used in the supervised RMOT loss. Target annotations are used only for evaluation.
	
	We use HOTA as the primary metric and DetA/AssA as the main diagnostic decomposition in the main paper. Unless otherwise specified, all methods are trained with AdamW for 60 epochs on 4 NVIDIA RTX 3090 GPUs. QCA adds the objectives in Eq.~(\ref{eq:overall_loss}) on top of the source-domain supervised RMOT loss.
	
	\subsection{Main Results and Findings}
	
	The in-domain TempRMOT reference on \textit{vKITTI clear} $\rightarrow$ \textit{clear} reaches 28.69 HOTA, 17.38 DetA, and 47.39 AssA, serving as the reference for the controlled degradation shown in Fig.~\ref{fig1}. Changing only the target visual domain already causes clear drops under shared scene semantics, confirming that language-conditioned UDA for RMOT is challenging even before moving to real-domain transfer.
	
	\begin{table}[!htbp]
		\caption{Representative results on CD-RMOT-Bench. HOTA, DetA, and AssA are reported for weather, viewpoint, synthetic--real, and real--synthetic transfer. MUTR and ReferDINO are included as RVOS-style references.}
		\centering
		\scriptsize
		\renewcommand{\arraystretch}{1.08}
		\setlength{\tabcolsep}{3.2pt}
		
		\resizebox{0.8\linewidth}{!}{%
			\begin{tabular}{llccccccccc}
				\toprule
				\multirow{2}{*}{Family} & \multirow{2}{*}{Method}
				& \multicolumn{3}{c}{\makecell[c]{vKITTI clear\\$\rightarrow$ fog}}
				& \multicolumn{3}{c}{\makecell[c]{vKITTI clear\\$\rightarrow$ rain}}
				& \multicolumn{3}{c}{\makecell[c]{vKITTI clear\\$\rightarrow$ 30\degree L}} \\
				\cmidrule(lr){3-5} \cmidrule(lr){6-8} \cmidrule(lr){9-11}
				& & HOTA & DetA & AssA & HOTA & DetA & AssA & HOTA & DetA & AssA \\
				\midrule
				\multirow{2}{*}{RVOS}
				& MUTR~\cite{yan2024referred} & 4.81 & 3.25 & 7.11 & 4.76 & 3.81 & 5.95 & 6.34 & 3.55 & 11.34 \\
				& ReferDINO~\cite{liang2025referdino} & 5.88 & 4.19 & 8.25 & 5.58 & 4.41 & 7.07 & 6.78 & 4.03 & 11.43 \\
				\midrule
				\multirow{4}{*}{RMOT}
				& TransRMOT~\cite{wu2023rmot} & 11.67 & 4.10 & 33.95 & 11.91 & 3.95 & 36.67 & 9.99 & 3.06 & 33.53 \\
				& MGLT~\cite{chen2025multigranularity} & 10.37 & 4.09 & 26.36 & 12.40 & 4.49 & 34.39 & 9.73 & 2.47 & 38.58 \\
				& TempRMOT~\cite{zhang2024bootstrapping} & 16.00 & 8.79 & 29.20 & 19.50 & 10.30 & 37.05 & 20.43 & 10.08 & 41.79 \\
				& QCA (ours) & \textbf{21.37} & \textbf{11.04} & \textbf{41.38} & \textbf{24.20} & \textbf{13.48} & \textbf{43.49} & \textbf{25.25} & \textbf{14.43} & \textbf{44.31} \\
				\bottomrule
			\end{tabular}
		}
		
		\vspace{0.35em}
		
		\resizebox{0.8\linewidth}{!}{%
			\begin{tabular}{llccccccccc}
				\toprule
				\multirow{2}{*}{Family} & \multirow{2}{*}{Method}
				& \multicolumn{3}{c}{\makecell[c]{vKITTI clear\\$\rightarrow$ Refer-KITTI-V2}}
				& \multicolumn{3}{c}{\makecell[c]{vKITTI clear\\$\rightarrow$ Refer-BDD}}
				& \multicolumn{3}{c}{\makecell[c]{Refer-KITTI-V2\\$\rightarrow$ vKITTI fog}} \\
				\cmidrule(lr){3-5} \cmidrule(lr){6-8} \cmidrule(lr){9-11}
				& & HOTA & DetA & AssA & HOTA & DetA & AssA & HOTA & DetA & AssA \\
				\midrule
				\multirow{2}{*}{RVOS}
				& MUTR~\cite{yan2024referred} & 5.33 & 2.38 & 11.90 & 4.91 & 2.45 & 9.86 & 3.48 & 2.15 & 5.63 \\
				& ReferDINO~\cite{liang2025referdino} & 5.85 & 2.64 & 12.94 & 6.29 & 3.02 & 13.09 & 4.28 & 2.68 & 6.84 \\
				\midrule
				\multirow{4}{*}{RMOT}
				& TransRMOT~\cite{wu2023rmot} & 9.87 & 3.10 & 31.79 & 12.74 & 4.33 & 38.52 & 10.59 & 4.56 & 24.64 \\
				& MGLT~\cite{chen2025multigranularity} & 12.88 & 4.95 & 33.84 & 13.04 & 5.87 & 28.96 & 11.28 & 4.95 & 25.73 \\
				& TempRMOT~\cite{zhang2024bootstrapping} & 17.17 & 7.97 & 37.60 & 14.32 & 6.29 & 33.95 & 11.47 & \textbf{5.44} & 24.57 \\
				& QCA (ours) & \textbf{18.92} & \textbf{8.76} & \textbf{41.32} & \textbf{19.09} & \textbf{8.17} & \textbf{44.91} & \textbf{14.86} & 5.10 & \textbf{43.98} \\
				\bottomrule
			\end{tabular}
		}
		\label{tab:vclear_to_target_comparison}
	\end{table}
	
	Table~\ref{tab:vclear_to_target_comparison} reports representative results across weather, viewpoint, synthetic-to-real, and real-to-synthetic transfer. The adapted RVOS baselines remain far behind native RMOT methods under the trajectory protocol. Since RVOS predicts language-conditioned masks rather than identity-aware trajectories, this gap suggests that video-language segmentation does not directly solve CD-RMOT. Details of the pseudo-mask adaptation are provided in \textbf{Appendix~\ref{sec:appendix_rvos}}.
	
	Among the controlled vKITTI variants, \textit{fog} is the most difficult target for TempRMOT, reducing HOTA from 28.69 to 16.00 relative to the in-domain \textit{vKITTI clear} $\rightarrow$ \textit{clear} reference. The degradation is not confined to a single metric component: DetA also drops under domain shift, while AssA shows especially large degradation in fog and rain. This pattern indicates that controlled shifts affect both target localization under the RMOT protocol and expression-consistent temporal selection. It also supports the design of Refer-vKITTI as a controlled digital-twin benchmark core rather than a generic heterogeneous dataset collection.
	
	Across all reported transfer settings, QCA achieves the strongest HOTA and AssA among the evaluated methods. The gains are particularly informative because QCA does not use target-domain boxes, identities, masks, or expression-object associations as supervised labels; it improves transfer by stabilizing and refining language-conditioned queries. At the same time, the results remain below the in-domain reference, indicating that CD-RMOT-Bench exposes substantial remaining room for language-conditioned CD-RMOT rather than a saturated comparison.
	
	The synthetic--real and reverse directions further show that the difficulty is not limited to controlled synthetic variants. Both \textit{vKITTI clear} $\rightarrow$ Refer-BDD and \textit{Refer-KITTI-V2} $\rightarrow$ \textit{vKITTI fog} remain challenging, supporting the need to evaluate weather, viewpoint, and realism shifts within a unified protocol. Full metric decompositions are reported in \textbf{Appendix~\ref{sec:appendix_fullmetrics}}.
	
	\subsection{Ablation and Qualitative Analysis}
	
	Ablation results are reported in \textbf{Appendix~\ref{sec:appendix_ablation}}. The three QCA modules correspond to complementary failure modes exposed by CD-RMOT-Bench: CDT-MT improves target-query stability, ACDA reduces task-relevant source--target query mismatch using naturally available domain labels, and LGCA uses language as a semantic anchor to reduce expression-conditioned drift. The stronger two-module and full-model variants indicate that language-conditioned UDA for RMOT requires joint treatment of query stability, domain alignment, and language-guided correction.
	
	Qualitative cases are provided in \textbf{Appendix~\ref{sec:appendix_qual}}, further illustrating that visually plausible predictions can still drift to expression-inconsistent trajectories under domain shift.
	
	\vspace{-5pt}
	\section{Conclusion}
	
	We introduced CD-RMOT-Bench for language-conditioned UDA in RMOT, where target-domain expressions are available as task inputs while target tracking annotations are used only for evaluation. Built from Refer-KITTI-V2, Refer-vKITTI, and Refer-BDD, the benchmark supports controlled weather/viewpoint shifts and bidirectional synthetic--real transfer under a unified RMOT protocol. Refer-vKITTI is constructed through per-scene identity mapping, domain-specific annotation transfer, and residual quality control. Experiments show that cross-domain degradation involves not only detection-sensitive errors, but also failures in expression-consistent temporal association and target selection. QCA provides a strong query-centric baseline, while the remaining gaps highlight the need for robust language-guided tracking across visual domains.
	
	\clearpage
	
	\bibliographystyle{unsrtnat}
	\bibliography{ref_final}
	
	\clearpage
	\appendix

\section{Benchmark Construction and Protocols}
\label{sec:appendix_construction}

\subsection{Alignment, Matching, and Residual Calibration}
\label{sec:appendix_alignment}

Refer-vKITTI is constructed from the five Refer-KITTI-V2 sequences that have aligned Virtual KITTI 2 counterparts. For each paired real/synthetic scene, the construction pipeline loads the Refer-KITTI-V2 frame-level labels and the vKITTI bounding boxes, converts them into a common box representation, and enumerates source--target identity candidates after scene and frame alignment. This formulation turns expression transfer into a constrained identity-mapping problem rather than an unconstrained annotation task, which is essential for preserving referential semantics under controlled domain changes.

Candidate evidence is collected through frame-level matching with an IoU threshold of 0.02. This threshold is intentionally permissive because it is used to collect correspondence evidence before scene-level assignment, not as a standalone acceptance rule. For each source--target identity pair, the pipeline computes frame-match count, mean frame-level IoU, direct track IoU over overlapping frames, center similarity, area similarity, and temporal IoU. These terms define the per-scene matching score
\begin{equation}
	s(i,j)
	=
	4N_{ij}
	+
	30\bar{I}^{\text{frame}}_{ij}
	+
	6\bar{I}^{\text{track}}_{ij}
	+
	2C_{ij}
	+
	A_{ij}
	+
	3T_{ij},
	\label{eq:app_matching_score}
\end{equation}
where $N_{ij}$ is the number of IoU-filtered frame matches, $\bar{I}^{\text{frame}}_{ij}$ is the mean frame-level IoU, $\bar{I}^{\text{track}}_{ij}$ is the direct track IoU over overlapping frames, $C_{ij}$ and $A_{ij}$ denote center and area similarity, and $T_{ij}$ denotes temporal IoU. The resulting scene-level identity correspondences are selected under a one-to-one assignment constraint, yielding auditable per-scene KITTI-to-vKITTI identity mappings.

The resulting mappings are then used to transfer Refer-KITTI-V2 expression-track annotations to all vKITTI domains. During normalization, the natural-language expression is preserved, video identifiers are rewritten to the corresponding vKITTI domain, target identities are remapped according to the per-scene correspondence, and frame-level targets are removed when the mapped vKITTI identity is absent in the target-domain frame. This accounts for visibility changes, viewpoint-induced invalidation, and domain-specific object absence, while keeping the final annotations compatible with the unified RMOT evaluation protocol.

\begin{table}[!htbp]
	\caption{Detailed construction statistics for Refer-vKITTI. Expression--track pairs and per-scene identity pairs are different counting units: the former count expression--track instances after domain expansion, while the latter count candidate identity correspondences used during scene-level matching. The automatic stage provides an auditable initial mapping; manual inspection is used only as residual quality control and is therefore described in the text rather than reported as a separate logged statistic.}
	\centering
	\scriptsize
	\setlength{\tabcolsep}{3.2pt}
	\renewcommand{\arraystretch}{1.08}
	\resizebox{0.9\textwidth}{!}{%
		\begin{tabular}{lp{2.8cm}p{6.6cm}}
			\toprule
			Item & Value & Evidence / Notes \\
			\midrule
			Total transferred videos & 50 & Five aligned KITTI/vKITTI scenes expanded to 10 vKITTI domains \\
			Total transferred expressions & 27{,}730 & Final Refer-vKITTI expression directory \\
			Total transferred expression--track pairs & 285{,}160 & Source expression--track pairs expanded across 10 target domains \\
			\midrule
			Candidate identity pairs after alignment & 6{,}887 & Per-scene source--target identity pairs with temporal frame overlap \\
			Candidate pairs with IoU evidence & 374 & Frame-level candidate evidence collected with IoU $\geq 0.02$ \\
			Matched identity pairs after one-to-one assignment & 251 & Per-scene identity mappings summed over five aligned scenes \\
			Unmatched source identities in build summary & 0 & Reported by scene-level mapping summaries \\
			\midrule
			Frame-level visibility invalidations & 806{,}110 & Dropped target mentions when mapped vKITTI identities are absent in target-domain frames \\
			Discarded expression--track pairs & 13{,}117 & Initial transferred pairs minus final retained pairs \\
			Final retained expression--track pairs & 272{,}043 & Expression--track pairs retained after visibility filtering and normalization \\
			Final retention rate & 95.40\% & Final retained pairs divided by initially transferred pairs \\
			\midrule
			IoU threshold for evidence collection & 0.02 & Used before scene-level assignment, not as a standalone acceptance rule \\
			Per-scene matching score & Composite score & Frame-match count, frame IoU, direct track IoU, center similarity, area similarity, and temporal IoU \\
			Temporal consistency cue & Soft score term & Temporal IoU contributes to the matching score \\
			Lightweight consistency check & 306{,}505 checks, 0 flagged & Post-construction consistency validation on generated labels \\
			\bottomrule
		\end{tabular}%
	}
	\label{tab:construction_quality_appendix}
\end{table}

Manual quality control is used only as a residual calibration step after automatic matching and transfer. We inspect ambiguous cases with weak correspondence evidence, short temporal overlap, frequent target disappearance, or dense same-class distractors. Each inspected case is checked against the source frame, the target-domain frame, the referring expression, and the mapped target identity. When necessary, identity assignments are corrected, invisible frame-level targets are removed, or expression--target pairs are invalidated when their semantics no longer hold after domain transfer. This keeps the main construction process automatic and auditable while allowing limited human calibration for cases where correspondence evidence is ambiguous. The procedure is used for quality control rather than large-scale manual annotation. Table~\ref{tab:construction_quality_appendix} summarizes the scale, identity-matching evidence, retained annotations, and post-construction consistency checks of the Refer-vKITTI construction pipeline.

\subsection{Splits, Protocols, and Implementation}
\label{sec:appendix_protocol}

For controlled vKITTI transfer, we use three clear-domain Refer-vKITTI scenes for source training, containing 1{,}407 frames and 1{,}710 expressions. Each reported target setting, including \textit{fog}, \textit{rain}, and \textit{30\degree L}, uses two held-out aligned scenes with 659 frames and 1{,}063 expressions. The real clear-domain setting follows the same cross-scene principle in Refer-KITTI-V2, using three videos for training with 1{,}359 frames and 1{,}710 expressions and two held-out videos for evaluation with 646 frames and 1{,}063 expressions. For synthetic-to-real transfer, Refer-BDD provides 8{,}385 training frames and 3{,}937 expressions for adaptation, and 1{,}619 test frames with 673 expressions for evaluation.

Refer-vKITTI is designed as a controlled digital-twin domain-shift protocol rather than a scene-disjoint or identity-disjoint generalization benchmark. Its purpose is to isolate weather and viewpoint changes under shared scene geometry and transferred language annotations. We complement this controlled setting with synthetic-to-real and real-to-synthetic experiments involving Refer-KITTI-V2 and Refer-BDD.

\begin{table}[!htbp]
	\caption{Clarification of the language-conditioned UDA setting used in CD-RMOT-Bench. Target-domain referring expressions are available as task inputs, but target-domain boxes, identities, masks, and expression-object associations are not used as supervised training labels.}
	\centering
	\scriptsize
	\setlength{\tabcolsep}{4pt}
	\renewcommand{\arraystretch}{1.10}
	\begin{tabular}{lccc}
		\toprule
		Component & Source training & Target adaptation & Evaluation \\
		\midrule
		Videos & \checkmark & \checkmark & \checkmark \\
		Referring expressions & \checkmark & \checkmark\; task inputs & \checkmark \\
		Bounding boxes & \checkmark & -- & \checkmark \\
		Track identities & \checkmark & -- & \checkmark \\
		Masks / pseudo masks & optional & -- & optional \\
		Expression-object associations & \checkmark & -- & \checkmark \\
		Supervised RMOT loss & \checkmark & -- & -- \\
		Prediction-level consistency & -- & \checkmark & -- \\
		Source--target domain alignment & \checkmark & \checkmark & -- \\
		\bottomrule
	\end{tabular}
	\label{tab:uda_setting_clarification}
\end{table}

All experiments follow language-conditioned UDA for RMOT. Source-domain videos, referring expressions, and tracking annotations are used for supervised RMOT training. Target-domain videos and referring expressions are used as task-conditioned inputs during adaptation, but target-domain bounding boxes, track identities, masks, and expression-object associations are not used as supervised training labels. Target annotations are reserved for evaluation. Table~\ref{tab:uda_setting_clarification} summarizes this distinction.

Unless otherwise specified, all methods are trained under the same optimization and evaluation setting. We use AdamW for 60 epochs, with the learning rate decayed after epoch 40, and batch size 1 per GPU on 4 NVIDIA RTX 3090 GPUs. For QCA, we use EMA decay $\alpha=0.999$ in CDT-MT, retain top-$M=16$ query features in ACDA, and set prototype momentum $m=0.9$ in LGCA. Additional schedule details are provided in Appendix~\ref{sec:appendix_method}.

\section{Reference Method Details}
\label{sec:appendix_method}

\subsection{CDT-MT: Reliable Query Selection}
\label{sec:appendix_cdtmt}

The main paper summarizes CDT-MT as selective temporal self-distillation over reliable target-domain queries. Here we provide the reliability estimation and subset selection details.

For target query $i$, we compute the teacher reliability score
\begin{equation}
	\rho_i = \sigma(o_i^{\text{tea}})\cdot \sigma(r_i^{\text{tea}}),
	\label{eq:app_mt_reliability}
\end{equation}
which measures whether the teacher regards the query as both object-like and expression-relevant.

We further quantify student--teacher agreement in three complementary spaces:
\begin{equation}
	a_i^{\text{feat}} = \frac{1+\cos(q_i^{\text{stu}},q_i^{\text{tea}})}{2},
	\label{eq:app_mt_feat_agree}
\end{equation}
\begin{equation}
	a_i^{\text{box}} = \exp\!\left(-4\|b_i^{\text{stu}}-b_i^{\text{tea}}\|_1\right),
	\label{eq:app_mt_box_agree}
\end{equation}
\begin{equation}
	a_i^{\text{ref}} = 1-\left|\sigma(r_i^{\text{stu}})-\sigma(r_i^{\text{tea}})\right|,
	\label{eq:app_mt_ref_agree}
\end{equation}
and combine them as
\begin{equation}
	a_i = \frac{a_i^{\text{feat}}+a_i^{\text{box}}+a_i^{\text{ref}}}{3}.
	\label{eq:app_mt_agree}
\end{equation}

The final selection score is
\begin{equation}
	\psi_i = (1-\gamma)\rho_i + \gamma a_i,
	\label{eq:app_mt_select_score}
\end{equation}
where $\gamma\in[0,1]$ balances teacher reliability and cross-model agreement. Queries with high $\psi_i$ are retained in $\Omega_t$ for consistency regularization. To avoid over-regularizing early training, we apply the warmup factor
\begin{equation}
	\mathcal{L}_{\text{mt}} \leftarrow \eta(e)\mathcal{L}_{\text{mt}},
	\qquad
	\eta(e)=\min\left(1,\frac{e+1}{E_w}\right).
	\label{eq:app_mt_warmup}
\end{equation}

\subsection{ACDA: Adversarial Schedule and Curriculum}
\label{sec:appendix_acda}

ACDA performs query-level adversarial alignment over task-relevant object queries. The domain supervision does not require additional manual annotation: each sample naturally belongs to the source or target split, so the split identity provides the domain signal used by the discriminator.

Let
\begin{equation}
	\tau_e=\frac{\max(0,e-E_w)}{\max(1,E-E_w)},
	\label{eq:app_acda_progress}
\end{equation}
denote normalized training progress after warmup, where $E$ is the total number of epochs. The adversarial strength is scheduled as
\begin{equation}
	\lambda_{\text{grl}}(e)
	=
	\lambda_{\min}
	+
	(\lambda_{\max}-\lambda_{\min})
	\left(
	\frac{2}{1+\exp(-10\tau_e)}-1
	\right).
	\label{eq:app_grl_schedule}
\end{equation}

To avoid forcing severely degraded target queries into early adversarial matching, ACDA introduces a curriculum driven by three signals: domain uncertainty, distance to a running source prototype, and RMOT task difficulty. Let $\pi_i=g(z_i)$ denote the domain posterior over $K_d$ domains, and let $\mu_s$ denote the running source prototype. We compute
\begin{equation}
	h_i^{\text{ent}}
	=
	-\frac{1}{\log K_d}
	\sum_{k=1}^{K_d}
	\pi_{ik}\log \pi_{ik},
	\label{eq:app_acda_entropy}
\end{equation}
\begin{equation}
	h_i^{\text{proto}}
	=
	\frac{\|z_i-\mu_s\|_2-\min_j\|z_j-\mu_s\|_2}
	{\max_j\|z_j-\mu_s\|_2-\min_j\|z_j-\mu_s\|_2+\varepsilon},
	\label{eq:app_acda_proto}
\end{equation}
\begin{equation}
	h_i^{\text{task}} = 1-\widetilde{c}_i,
	\label{eq:app_acda_taskdiff}
\end{equation}
where $\widetilde{c}_i$ is the normalized task confidence. The final difficulty score is
\begin{equation}
	h_i = 0.5\,h_i^{\text{ent}} + 0.25\,h_i^{\text{proto}} + 0.25\,h_i^{\text{task}},
	\label{eq:app_acda_difficulty}
\end{equation}
and the resulting curriculum weight is
\begin{equation}
	w_i^{\text{acda}}
	=
	\omega_{\min}
	+
	(1-\omega_{\min})
	\Bigl(
	\tau_e + (1-\tau_e)(1-h_i)
	\Bigr).
	\label{eq:app_acda_weight}
\end{equation}

\subsection{LGCA: Fusion and Auxiliary Objectives}
\label{sec:appendix_lgca}

LGCA uses the referring expression as a semantic anchor for prototype-based query correction. The main paper provides the language-guided confidence, prototype update, and residual refinement. Here we include the omitted fusion rule and auxiliary training terms.

LGCA decomposes the corrected representation into appearance-oriented and motion-oriented anchors, producing $q_i^{\text{app}}$ and $q_i^{\text{mot}}$. Their combination is controlled by a quality-aware gate:
\begin{equation}
	q_i^{*} = g_a(\zeta)\,q_i^{\text{app}} + g_m(\zeta)\,q_i^{\text{mot}},
	\qquad
	g_a(\zeta)+g_m(\zeta)=1,
	\label{eq:app_lgca_fusion}
\end{equation}
where $\zeta$ denotes sequence quality estimated from prototype-query agreement.

During training, LGCA uses the compact objective
\begin{equation}
	\mathcal{L}_{\text{lgca}}
	=
	\mathcal{L}_{\text{align}}
	+
	\lambda_{\text{cons}}\mathcal{L}_{\text{cons}}
	+
	\lambda_{\text{qual}}\mathcal{L}_{\text{qual}}
	+
	\lambda_{\text{gate}}\mathcal{L}_{\text{gate}},
	\label{eq:app_lgca_loss}
\end{equation}
with
\begin{equation}
	\mathcal{L}_{\text{align}} = 1 - \frac{1}{N}\sum_i \cos(q_i^{*}, \ell),
	\label{eq:app_lgca_align}
\end{equation}
\begin{equation}
	\mathcal{L}_{\text{cons}} = \frac{1}{N}\sum_i \|q_i^{*}-q_i\|_2^2.
	\label{eq:app_lgca_cons}
\end{equation}
The remaining terms regularize prototype quality and gate behavior. During inference, the corrected representation is injected through the residual blend in Eq.~(\ref{eq:lgca_blend}) of the main paper.

\section{Additional Experimental Results}
\label{sec:appendix_exp}

\subsection{Adapted RVOS Baselines}
\label{sec:appendix_rvos}

RVOS methods predict language-conditioned mask sequences rather than RMOT trajectories. To include them as cross-task references, we adapt MUTR and ReferDINO through a pseudo-mask interface. For each video-expression pair, the RMOT bounding boxes associated with the referred track identities are rasterized into rectangular pseudo masks, yielding RVOS-style training samples without introducing extra mask annotation. The RVOS models are trained under the same source-domain splits and applied to the corresponding target-domain videos.

At inference time, predicted masks are thresholded and converted back into bounding boxes by taking the tight enclosing rectangle of the foreground region in each frame. Empty predictions are treated as missed detections. When multiple foreground components are present, we use the tight bounding box enclosing all foreground pixels. The resulting predictions are written into the same TrackEval-compatible trajectory format as native RMOT methods. These baselines should be interpreted as adapted comparison references rather than native RMOT trackers, because their association behavior is induced by per-expression mask-to-box conversion rather than explicit multi-object identity management. This adaptation is used only to provide additional cross-task reference points rather than direct competition with native RMOT methods.

\subsection{Extended Metric Decompositions}
\label{sec:appendix_fullmetrics}

\begin{table}[!htbp]
	\caption{Extended cross-domain metric decomposition under the unified evaluation protocol. Values in parentheses denote differences relative to the TempRMOT baseline.}
	\centering
	\scriptsize
	\setlength{\tabcolsep}{3.8pt}
	\renewcommand{\arraystretch}{1.20}
	\resizebox{0.85\linewidth}{!}{%
		\begin{tabular}{llcccccccc}
			\toprule
			\multirow{2}{*}{\makecell[c]{Setting}} & \multirow{2}{*}{\makecell[c]{Method}} & \multirow{2}{*}{\makecell[c]{HOTA}} & \multicolumn{3}{c}{Detection} & \multicolumn{3}{c}{Association} & \multirow{2}{*}{\makecell[c]{LocA}} \\
			\cmidrule(lr){4-6}\cmidrule(lr){7-9}
			& & & \makecell[c]{DetA} & \makecell[c]{DetRe} & \makecell[c]{DetPr} & \makecell[c]{AssA} & \makecell[c]{AssRe} & \makecell[c]{AssPr} & \\
			\midrule
			
			\multirow{2}{*}[-0.6em]{\makecell[c]{vKITTI clear\\$\rightarrow$ fog}}
			& \makecell[c]{TempRMOT}
			& 16.00 & 8.79 & 14.20 & 18.58 & 29.20 & 34.45 & 70.61 & \textbf{88.87} \\
			& \makecell[c]{QCA}
			& \makecell[c]{21.37\\(+5.37)}
			& \makecell[c]{11.04\\(+2.25)}
			& \makecell[c]{18.02\\(+3.82)}
			& \makecell[c]{21.80\\(+3.22)}
			& \makecell[c]{41.38\\(+12.18)}
			& \makecell[c]{47.86\\(+13.41)}
			& \makecell[c]{72.10\\(+1.49)}
			& \makecell[c]{87.14\\(-1.73)} \\
			\midrule
			
			\multirow{2}{*}[-0.6em]{\makecell[c]{vKITTI clear\\$\rightarrow$ rain}}
			& \makecell[c]{TempRMOT}
			& 19.50 & 10.30 & 20.95 & 16.64 & 37.05 & 43.49 & 68.89 & 86.39 \\
			& \makecell[c]{QCA}
			& \makecell[c]{24.20\\(+4.70)}
			& \makecell[c]{13.48\\(+3.18)}
			& \makecell[c]{16.44\\(-4.51)}
			& \makecell[c]{41.93\\(+25.29)}
			& \makecell[c]{43.49\\(+6.44)}
			& \makecell[c]{51.04\\(+7.55)}
			& \makecell[c]{71.11\\(+2.22)}
			& \makecell[c]{89.90\\(+3.51)} \\
			\midrule
			
			\multirow{2}{*}[-0.6em]{\makecell[c]{vKITTI clear\\$\rightarrow$ 30\degree L}}
			& \makecell[c]{TempRMOT}
			& 20.43 & 10.08 & 23.92 & 14.65 & 41.79 & 53.74 & 67.46 & 85.45 \\
			& \makecell[c]{QCA}
			& \makecell[c]{25.25\\(+4.82)}
			& \makecell[c]{14.43\\(+4.35)}
			& \makecell[c]{19.85\\(-4.07)}
			& \makecell[c]{33.77\\(+19.12)}
			& \makecell[c]{44.31\\(+2.52)}
			& \makecell[c]{49.76\\(-3.98)}
			& \makecell[c]{78.42\\(+10.96)}
			& \makecell[c]{88.21\\(+2.76)} \\
			\midrule
			
			\multirow{2}{*}[-0.6em]{\makecell[c]{vKITTI clear\\$\rightarrow$ Refer-KITTI-V2}}
			& \makecell[c]{TempRMOT}
			& 17.17 & 7.97 & 21.33 & 11.03 & 37.60 & 47.52 & 61.29 & 78.76 \\
			& \makecell[c]{QCA}
			& \makecell[c]{18.92\\(+1.75)}
			& \makecell[c]{8.76\\(+0.79)}
			& \makecell[c]{15.63\\(-5.70)}
			& \makecell[c]{16.11\\(+5.08)}
			& \makecell[c]{41.32\\(+3.72)}
			& \makecell[c]{49.19\\(+1.67)}
			& \makecell[c]{67.68\\(+6.39)}
			& \makecell[c]{79.28\\(+0.52)} \\
			\midrule
			
			\multirow{2}{*}[-0.6em]{\makecell[c]{vKITTI clear\\$\rightarrow$ Refer-BDD}}
			& \makecell[c]{TempRMOT}
			& 14.32 & 6.29 & 10.88 & 12.61 & 33.95 & 41.19 & 62.45 & 77.16 \\
			& \makecell[c]{QCA}
			& \makecell[c]{19.09\\(+4.77)}
			& \makecell[c]{8.17\\(+1.88)}
			& \makecell[c]{9.13\\(-1.75)}
			& \makecell[c]{42.83\\(+30.22)}
			& \makecell[c]{44.91\\(+10.96)}
			& \makecell[c]{51.84\\(+10.65)}
			& \makecell[c]{72.68\\(+10.23)}
			& \makecell[c]{88.61\\(+11.45)} \\
			\midrule
			
			\multirow{2}{*}[-0.6em]{\makecell[c]{Refer-KITTI-V2\\$\rightarrow$ vKITTI fog}}
			& \makecell[c]{TempRMOT}
			& 11.47 & 5.44 & 8.46 & 13.00 & 24.57 & 28.98 & 62.66 & 82.40 \\
			& \makecell[c]{QCA}
			& \makecell[c]{14.86\\(+3.39)}
			& \makecell[c]{5.10\\(-0.34)}
			& \makecell[c]{16.18\\(+7.72)}
			& \makecell[c]{6.83\\(-6.17)}
			& \makecell[c]{43.98\\(+19.41)}
			& \makecell[c]{54.74\\(+25.76)}
			& \makecell[c]{63.00\\(+0.34)}
			& \makecell[c]{78.45\\(-3.95)} \\
			\bottomrule
		\end{tabular}%
	}
	\label{tab:full_metric_main}
\end{table}

Table~\ref{tab:full_metric_main} reports extended metric decompositions for TempRMOT and QCA under the reported transfer settings. The main paper focuses on HOTA, DetA, and AssA, while this appendix further reports detection, association, and localization components to support the diagnostic analysis.

\subsection{Ablation Details}
\label{sec:appendix_ablation}

\begin{table}[!htbp]
	\caption{Ablation study under two representative cross-domain transfer directions: vKITTI clear $\rightarrow$ fog and vKITTI clear $\rightarrow$ Refer-BDD. HOTA, DetA, and AssA are reported to show how each component affects detection- and association-sensitive performance.}
	\centering
	\scriptsize
	\setlength{\tabcolsep}{3.5pt}
	\renewcommand{\arraystretch}{1.08}
	\resizebox{0.85\linewidth}{!}{%
		\begin{tabular}{lccc ccc ccc}
			\toprule
			\multirow{2}{*}{Method} & \multirow{2}{*}{CDT-MT} & \multirow{2}{*}{ACDA} & \multirow{2}{*}{LGCA}
			& \multicolumn{3}{c}{vKITTI clear $\rightarrow$ fog}
			& \multicolumn{3}{c}{vKITTI clear $\rightarrow$ Refer-BDD} \\
			\cmidrule(lr){5-7} \cmidrule(lr){8-10}
			& & & & HOTA & DetA & AssA & HOTA & DetA & AssA \\
			\midrule
			TempRMOT baseline &  &  &  & 16.00 & 8.79 & 29.20 & 14.32 & 6.29 & 33.95 \\
			LGCA only &  &  & \checkmark & 16.17 & 8.62 & 30.34 & 15.14 & 6.21 & 36.80 \\
			ACDA only &  & \checkmark &  & 16.34 & 8.25 & 32.36 & 14.79 & 5.90 & 37.05 \\
			CDT-MT only & \checkmark &  &  & 16.25 & 8.48 & 31.14 & 14.61 & 6.03 & 35.35 \\
			CDT-MT + ACDA & \checkmark & \checkmark &  & 16.58 & 7.92 & 34.89 & 15.69 & 6.34 & 38.83 \\
			CDT-MT + LGCA & \checkmark &  & \checkmark & 17.59 & 8.19 & 38.02 & 16.73 & 6.92 & 40.50 \\
			ACDA + LGCA &  & \checkmark & \checkmark & 18.73 & 8.98 & 39.08 & 17.84 & 7.46 & 42.70 \\
			CDT-MT + ACDA + LGCA & \checkmark & \checkmark & \checkmark & \textbf{21.37} & \textbf{11.04} & \textbf{41.38} & \textbf{19.09} & \textbf{8.17} & \textbf{44.91} \\
			\bottomrule
		\end{tabular}
	}
	\label{tab:ablation_main_app}
\end{table}

Table~\ref{tab:ablation_main_app} reports ablations under two representative transfer directions. The three components target complementary failure modes exposed by CD-RMOT-Bench: unreliable target queries, task-relevant source--target query mismatch, and language-conditioned semantic drift. Their combination gives the strongest overall performance, supporting the interpretation that language-conditioned UDA for RMOT requires joint treatment of query stability, domain alignment, and language-guided correction.

\subsection{Additional Qualitative Cases}
\label{sec:appendix_qual}

\begin{figure}[!htbp]
	\centering
	\includegraphics[width=0.8\linewidth]{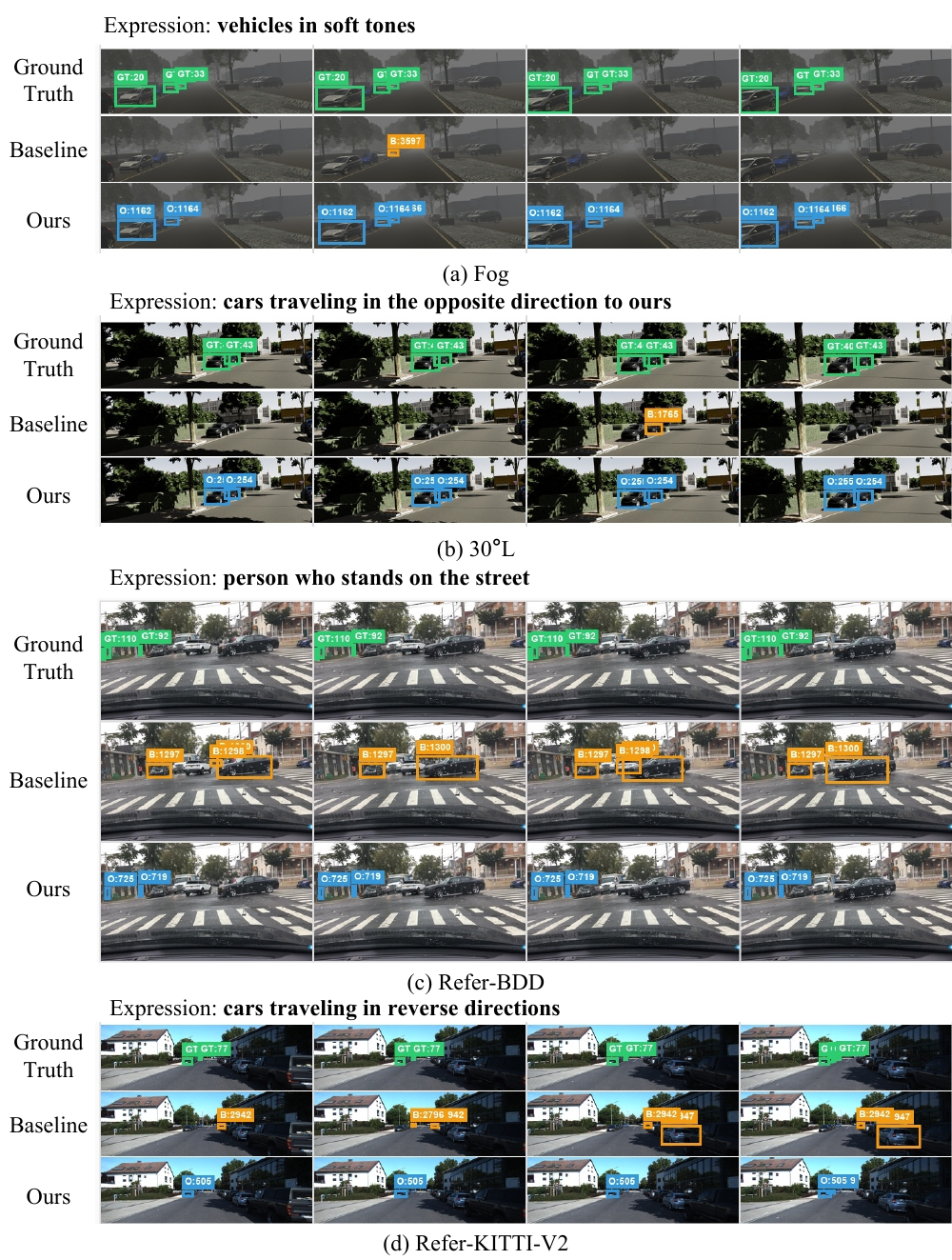}
	\caption{Additional qualitative cases illustrating common cross-domain RMOT failures. The examples show that domain shift can induce both distractor confusion and weakened expression-consistent temporal selectivity, even when visually plausible targets remain detectable.}
	\label{fig:qual_appendix}
\end{figure}

Fig.~\ref{fig:qual_appendix} provides qualitative comparisons under representative weather, viewpoint, real-domain, and motion-dependent settings. These cases support the diagnostic finding in the main paper: under domain shift, plausible objects may remain detectable while expression-consistent target selection drifts over time.

\end{document}